\documentclass[11pt,a4paper]{article}
\usepackage[latin9]{inputenc}
\usepackage{color}
\usepackage{array}
\usepackage{booktabs}
\usepackage{multirow}
\usepackage{amsmath}
\usepackage{graphicx}
\usepackage[authoryear]{natbib}
\usepackage{subscript}
\usepackage[unicode=true,pdfusetitle,
 bookmarks=true,bookmarksnumbered=false,bookmarksopen=false,
 breaklinks=false,pdfborder={0 0 1},backref=false,colorlinks=true]
 {hyperref}

\makeatletter

\pdfpageheight\paperheight
\pdfpagewidth\paperwidth

\providecommand{\tabularnewline}{\\}

\@ifundefined{date}{}{\date{}}
\usepackage{emnlp2018}
\usepackage{times}
\usepackage{latexsym}

\usepackage{url}

\aclfinalcopy 

\title{Amobee at IEST 2018: Transfer Learning from Language Models}

\author{Alon Rozental\thanks{~~These authors contributed equally to this work.}~~, Daniel Fleischer\footnotemark[1]~~, Zohar Kelrich\footnotemark[1] \\
  Amobee Inc., Tel Aviv, Israel \\
   \tt alon.rozental@amobee.com \\ \tt daniel.fleischer@amobee.com \\ \tt zohar.kelrich@amobee.com}

\makeatother

\begin{document}
\maketitle 
\begin{abstract}
This paper describes the system developed at Amobee for the WASSA
2018 implicit emotions shared task (IEST). The goal of this task was
to predict the emotion expressed by missing words in tweets without
an explicit mention of those words. We developed an ensemble system
consisting of language models together with LSTM-based networks containing
a CNN attention mechanism. Our approach represents a novel use of
language models\textemdash specifically trained on a large Twitter
dataset\textemdash to predict and classify emotions. Our system reached
1st place with a macro F\textsubscript{1} score of 0.7145.
\end{abstract}

\section{Introduction}

Sentiment analysis (SA) is a sub-field of natural language processing
(NLP) that explores the automatic deduction of feelings and attitudes
from textual data. One popular choice of source to study is Twitter,
a social network website where people publish short messages, called
tweets, with a maximum length of 280 characters. People write on various
topics, including global and local events, public figures, brands
and products. Twitter data has attracted the interest of both academia
and industry for the last several years. It contains some unique features,
such as emojis, misspelling and slang that are of interest to NLP
researchers while also containing insights relevant for business intelligence,
marketing and e-governance.

The implicit emotions \href{http://implicitemotions.wassa2018.com/}{shared task}
(IEST) is part of the \href{https://wt-public.emm4u.eu/wassa2018/}{WASSA 2018}
workshop, and is concerned with classifying tweets into one of 6 emotions\textemdash anger,
disgust, fear, joy, sadness and surprise\textemdash without an explicit
mention of emotion words. There were 30 teams who participated in
the task; for a description and analysis of the task and the datasets,
see \citet{Klinger2018x}.

This paper describes our specially developed system for the shared
task; it comprises several ensembles, where our new contribution is
the use of a language model as an emotion classifier. A language model,
based on the Transformer-Decoder architecture \citep{DBLP:journals/corr/VaswaniSPUJGKP17}
was trained using a large Twitter dataset, and used to produce probabilities
for each of the 6 emotions.

The paper is organized as follows: Sections \ref{sec:Data-Sources}
and \ref{sub:Embeddings-Training} describe our data sources and the
embedding training, Section \ref{sec:Language-Models} describes the
training and usage of the language models. In Section \ref{sec:Features}
we describe the resources that are used as features; Section \ref{sec:Building Blocks}
describes the architecture, broken into smaller components. Finally,
we review and conclude in Section \ref{sec:Summary}. 

\section{Data Sources\label{sec:Data-Sources}}

We used several data sources for the shared task:
\begin{enumerate}
\item Twitter Firehose: we took a random sample of 5 billion unique tweets
using the Twitter Firehose service. The tweets were used to train
language models and word embeddings; in the following, we will refer
to this as the Tweets\_5B dataset. 
\item Semeval 2018 shared task 1 datasets, specifically subtasks 1 and 5
in which tweets are classified into one of 4 emotions (anger, fear,
joy and sadness; subtask 1) and a multi-label classification of tweets
into 11 emotions (sub-task 5). We used both the datasets and our trained
models; \citet{Rozental:2018aa} describes the system and \citet{SemEval2018Task1}
describes the shared task.
\item The official IEST 2018 task datasets; the missing emotion words are
replaced by the keyword \texttt{{[}\#TRIGGERWORD\#{]}}. Table 
\begin{table}
\begin{centering}
\begin{tabular}{lccc}
\toprule 
\multicolumn{1}{c}{{\footnotesize{}Label}} & {\footnotesize{}Train} & {\footnotesize{}Dev} & {\footnotesize{}Test}\tabularnewline
\cmidrule[0.05em](rl){1-1}\cmidrule[0.05em](rl){2-2}\cmidrule[0.05em](rl){3-3}\cmidrule[0.05em](rl){4-4}{\footnotesize{}Anger} & {\footnotesize{}$25562$} & {\footnotesize{}$1600$} & {\footnotesize{}$4794$}\tabularnewline
{\footnotesize{}Disgust} & {\footnotesize{}$25558$} & {\footnotesize{}$1597$} & {\footnotesize{}$4794$}\tabularnewline
{\footnotesize{}Fear} & {\footnotesize{}$25575$} & {\footnotesize{}$1598$} & {\footnotesize{}$4791$}\tabularnewline
{\footnotesize{}Joy} & {\footnotesize{}$27958$} & {\footnotesize{}$1736$} & {\footnotesize{}$5246$}\tabularnewline
{\footnotesize{}Sad} & {\footnotesize{}$23165$} & {\footnotesize{}$1460$} & {\footnotesize{}$4340$}\tabularnewline
{\footnotesize{}Surprise} & {\footnotesize{}$25565$} & {\footnotesize{}$1600$} & {\footnotesize{}$4792$}\tabularnewline
\midrule 
\multicolumn{1}{c}{{\footnotesize{}Total}} & {\footnotesize{}$153383$} & {\footnotesize{}$9591$} & {\footnotesize{}$28757$}\tabularnewline
\bottomrule
\end{tabular}
\par\end{centering}
\caption{\label{tab:Distribution-of-labels}Distributions of labels in the
train, dev and test datasets.}
\end{table}
\ref{tab:Distribution-of-labels} presents the label distributions;
refer to the task paper for a description of the dataset.
\end{enumerate}
We used different pre-processing procedures on the aforementioned
tweets for our different learning algorithms. Those procedures ranged
from no pre-processing at all (for language models), through a simple
cleanup (for word embeddings) to an extensive pre-processing, used
with our Semeval \citeyearpar{Rozental:2018aa} system to produce
predictions, with the following processing steps: word tokenization,
part of speech tagging, regex treatment, lemmatization, named entity
recognition, synonym replacement and word replacement using a wikipedia-based
dictionary.

\section{Embeddings Training\label{sub:Embeddings-Training}}

Word embeddings are a set of algorithms designed to encode a large
vocabulary using low-dimensional real vectors. Depending on the algorithm,
the vectors carry additional semantic information, and are used in
down-stream NLP tasks. We trained word embeddings specifically for
the task; first, starting with the Tweets\_5B dataset, we removed
exact duplicates. Then we used a regex process: URLs, emails and Twitter
usernames were replaced with special keywords. Next we removed tweets
by using a text similarity threshold\footnote{~Using the \texttt{SequenceMatcher} module in Python.}.
Finally, we replaced rare words with a special token; the criterion
was to have a vocabulary of 300K unique tokens in total. We used the
\href{https://radimrehurek.com/gensim/index.html}{Gensim} package
\citep{rehurek_lrec} to train 4 embeddings with sizes of 300, 500,
700 and 1000 with the Word2vec \citep{NIPS2013_5021} algorithm. Similarly,
we trained 4 embeddings using the FastText algorithm \citep{Bojanowski:2017aa}.
We found that for the purpose of downstream tasks, the Word2vec embeddings
outperformed the FastText embeddings for each of the 4 sizes. In addition,
the Word2vec embedding of size 1000 performed better than the others,
provided that the training set is large enough. The size of the IEST
2018 train set was sufficiently large for us to use that single word
embedding. The embeddings usage is described in the architecture section
\ref{sec:Building Blocks}.

\section{Language Models\label{sec:Language-Models}}

We trained a language model (LM) using the Transformer-Decoder architecture,
introduced in \citet{DBLP:journals/corr/VaswaniSPUJGKP17}. We used
the \href{https://github.com/tensorflow/tensor2tensor}{Tensor2Tensor}
library \citep{tensor2tensor} with the built-in \texttt{transformer-big}
parameter set, where we only set the tweet maximum length to be 64
tokens. The model was trained for 2 days using the Tweets\_5B dataset
on 8 Nvidia Tesla V100 GPUs. We will refer to this model as LM1. We
built a pipeline around the trained model, such that given a sentence,
its probability to be randomly generated by the model is returned.
For example, under LM1 the probability of the text ``\textit{I was
surprised to see you here}'' (S1) being generated is $\exp\left(-25.76\right)$
and the text ``\textit{I was afraid to see you here}'' (S2) has
a probability of $\exp\left(-27.86\right)$. One can then calculate
the conditional probability of having S1 given only S1 or S2 were
generated, with a resulting value of $0.89$. 

In order to use LM1 to predict the correct label for a tweet, we created
a list of possible words for each of the six emotions, presented in
appendix \ref{sec:Emotions-Lexicon}. For each tweet, we replaced
the trigger word with each of the words from the list and then selected
the most probable version of each emotion. The resulting 6 normalized
probabilities are considered to be the probabilities assigned by the
LM for the possible labels. See table 
\begin{table*}
\centering{}%
\begin{tabular}{ccccc}
\toprule 
{\small{}Emotion} & {\small{}Possible Tweet} & {\small{}Log Probability} & {\small{}Max} & {\small{}Final Probability}\tabularnewline
\midrule
\multirow{2}{*}{{\footnotesize{}Joy}} & {\footnotesize{}I'm }\textbf{\footnotesize{}happy}{\footnotesize{}
than you.} & {\footnotesize{}$-24.38$} & \multirow{2}{*}{{\footnotesize{}$-19.7$}} & \multirow{2}{*}{{\small{}$0.89995$}}\tabularnewline
 & {\footnotesize{}I'm }\textbf{\footnotesize{}happier}{\footnotesize{}
than you.} & {\footnotesize{}$-19.7$} &  & \tabularnewline
\cmidrule{2-3} 
\multirow{2}{*}{{\footnotesize{}Angry}} & {\footnotesize{}I'm }\textbf{\footnotesize{}angry}{\footnotesize{}
than you.} & {\footnotesize{}$-26.8$} & \multirow{2}{*}{{\footnotesize{}$-21.9$}} & \multirow{2}{*}{{\small{}$0.09972$}}\tabularnewline
 & {\footnotesize{}I'm }\textbf{\footnotesize{}angrier}{\footnotesize{}
than you.} & {\footnotesize{}$-21.9$} &  & \tabularnewline
\cmidrule{2-3} 
\multirow{2}{*}{{\footnotesize{}Surprise}} & {\footnotesize{}I'm }\textbf{\footnotesize{}surprise}{\footnotesize{}
than you.} & {\footnotesize{}$-27.6$} & \multirow{2}{*}{{\footnotesize{}$-27.6$}} & \multirow{2}{*}{{\small{}$0.00033$}}\tabularnewline
 & {\footnotesize{}I'm }\textbf{\footnotesize{}surprised}{\footnotesize{}
than you.} & {\footnotesize{}$-31.5$} &  & \tabularnewline
\bottomrule
\end{tabular}\caption{\label{tab:LM1}Probability calculation of the sentence ``I'm \#\{TRIGGERWORD\}
than you.'' with 3 emotions using the language models. Notice that
the sentences which are grammatically incorrect have much lower probabilities. }
\end{table*}
\ref{tab:LM1} for a more detailed example with 3 emotions.

In addition to LM1, we trained another language model, denoted by
LM2; it was generated by taking LM1 and continuing its training using
just the tweets of the shared task dataset, where the trigger word
was replaced by the most probable word (according to LM1 predictions)
in the emotional category matching the label. LM2 was trained for
a day using a single V100 GPU. The prediction procedure was the same
as for LM1. For the purposes of downstream analysis, the features
we extracted from these models are the final 6 probabilities $p_{i}\left(s\right)$,
the log probability to generate the most likely candidate tweet by
random\textemdash referred to as tweet complexity\textemdash given
by $\text{comp}\left(s\right)=\underset{w\in\text{W}}{\max}\log p_{w}\left(s\right)$,
where W is the set of possible replacement words and finally, for
each candidate tweet, its shifted log probability, given by $\log\tilde{p}_{w}\left(s\right)=\log p_{w}\left(s\right)-\text{comp}\left(s\right)$. 

\section{Features\label{sec:Features}}

We used 4 types of features in our system: first we used predictions
from the language models; we took both the log-probabilities of the
tweets with the trigger words replaced by each word from appendix
\ref{sec:Emotions-Lexicon}, as well as the final 6 probabilities
for each tweet, for each of the two language model. Next, we used
our system for the Semeval 2018 task 1 competition to generate features
and predictions for sub-tasks 1 and 5 (as mentioned in section \ref{sec:Data-Sources}).
Next we used 2 external resources for tweets embedding: Universal
Sentence Embedding \citep{Cer:2018aa}, using the \href{https://www.tensorflow.org/hub/modules/google/universal-sentence-encoder/2}{Tensorflow Hub}
service and the \href{https://deepmoji.mit.edu/}{DeepMoji} package
\citep{felbo2017}. We created 7 versions of each tweet by replacing
the trigger word with one of the 6 emotions and an unrecognized word,
thus creating 7 Universal Sentence Embedding of dimension 512. The
DeepMoji embedding size is 2304 and only one was produced for each
tweet. Finally, we added a binary feature that captures whether the
trigger word has a prefix in each tweet. These features are used in
the 1st \eqref{sec:MEGA} and 2nd \eqref{sec:Ensemble} ensembles. 

\section{Architecture Overview\label{sec:Building Blocks} }

The system comprises of a multi-level soft-voting ensemble. Each building
block described in this section is a classifier by itself and is presented
as such. For our submitted solution, the building blocks were trained
jointly in the manner described in the next section, using a single
Nvidia GTX 1080 Ti GPU. We used the \href{https://keras.io/}{Keras}
library \citep{chollet2015keras} and the \href{https://www.tensorflow.org/}{TensorFlow}
framework \citep{Abadi:2016aa}. 

\subsection{Mini ASC Modules\label{subsec:Mini-ASC-Modules}}

This component consists of a bi-LSTM layer with a CNN-based attention
mechanism, similar to a single module in the Amobee Sentiment Classifier
(ASC) architecture described in \citeyearpar{Rozental:2018aa}. A
Dropout layer \citep{Srivastava:2014aa} of $0.5$ was applied between
each 2 consecutive layers except for the word embedding layer; for
an illustration, see figure 
\begin{figure*}
\begin{centering}
\includegraphics[scale=0.25]{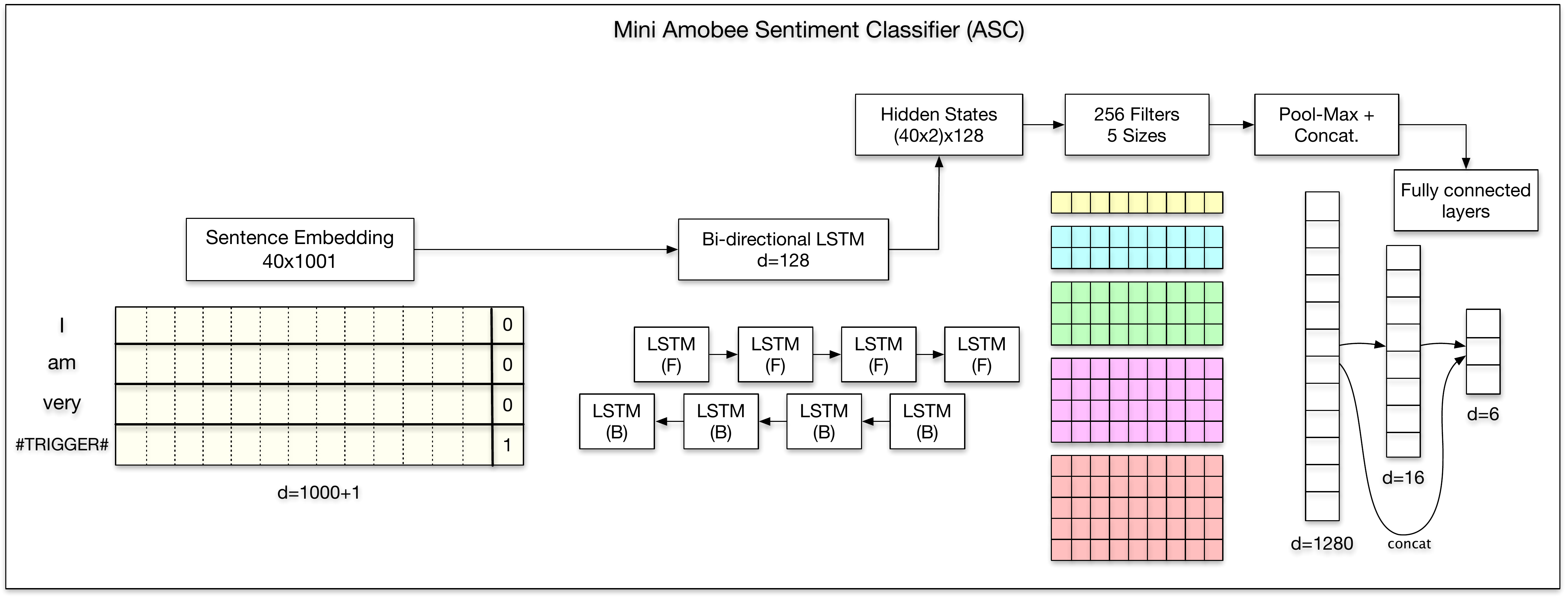}
\par\end{centering}
\caption{\label{fig:Mini-Model}Architecture of the mini Amobee sentiment classifier.}
\end{figure*}
\ref{fig:Mini-Model}. The input was the official dataset, transformed
using our trained embeddings, where the trigger word was embedded
as an unknown word using the rare-words token. We concatenated an
additional bit to each word vector, denoting whether it is the missing
trigger-word, differentiating it from other unknown words. There are
three key differences from our original work: 
\begin{enumerate}
\item The GRU layer was replaced by an LSTM layer.
\item Residual connections were added from the output of the max-pooling
layer to the network output.
\item Hyper parameters values were in the following ranges: embedding size=1000,
LSTM hidden size={[}128, 512{]}, number of filters={[}128, 512{]}
and dense layer size={[}16, 32{]}. 
\end{enumerate}
Training a single mini-ASC module on the IEST 2018 training set using
the Adam optimizer \citep{kingma2014adam}, categorial cross entropy
loss function and a batch size of 32, results in an average accuracy
of $0.669$ on the official validation set.

\subsection{First Ensemble\label{sec:MEGA}}

\begin{figure*}[t]
\centering{}\includegraphics[scale=0.42]{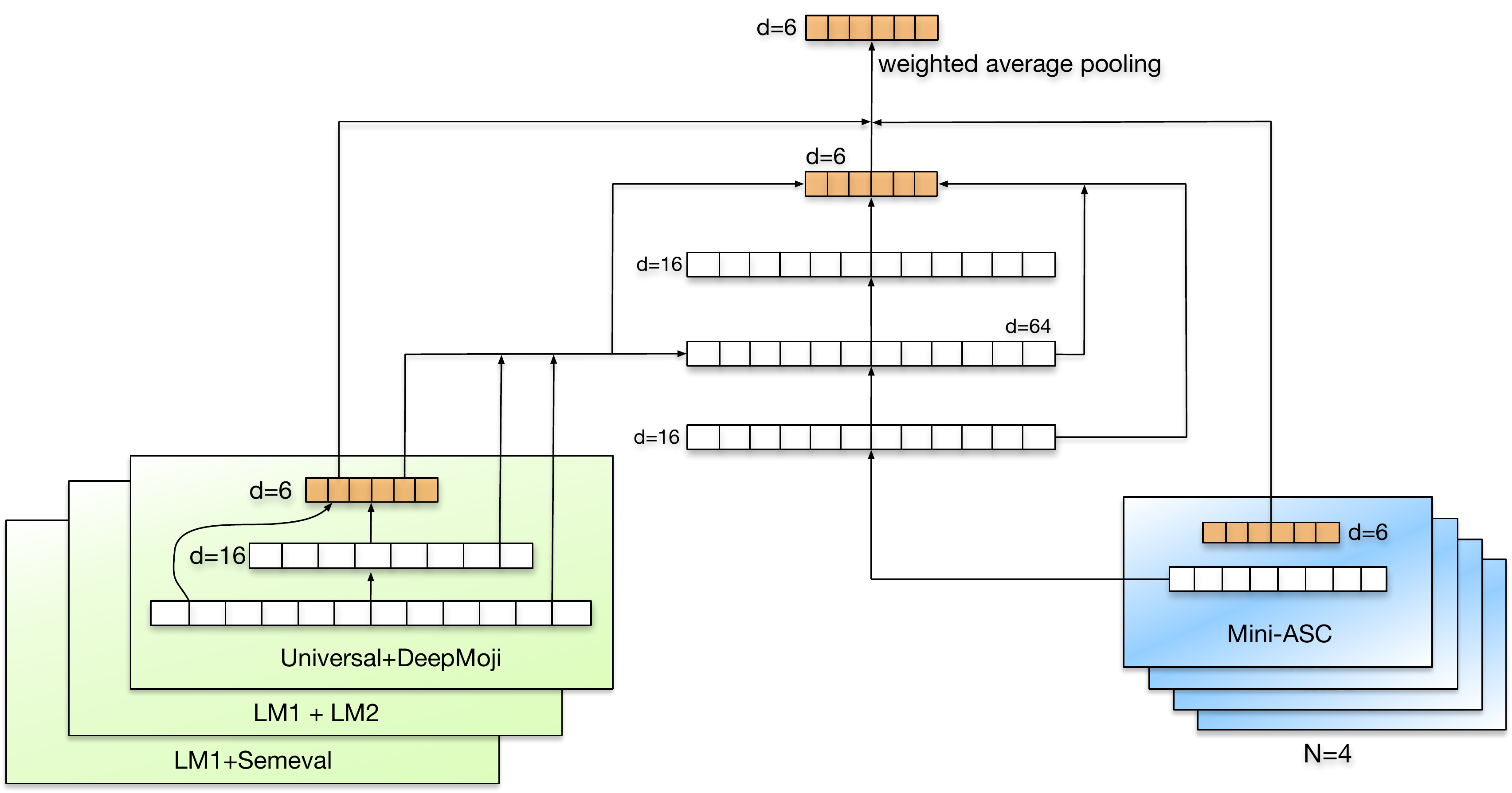}\caption{\label{fig:E1}Architecture of the first-level ensemble.}
\end{figure*}

The first level ensemble incorporates 4 mini ASC modules and 3 identical
sub-networks (see figure \ref{fig:E1}). The sub-networks share the
same architecture and their inputs are the following:
\begin{enumerate}
\item Universal + DeepMoji embeddings; this network reaches an average F\textsubscript{1}
score of 0.587 by itself on the validation set.
\item The LM1 + LM2 predictions; this network reaches an average F\textsubscript{1}
score of 0.637 by itself on the validation set.
\item The Semeval 2018 predictions, together with the LM1 predictions; this
network reaches an average F\textsubscript{1} score of 0.646 by itself
on the validation set.
\end{enumerate}
These networks share the same structure: the input is connected to
a dense layer of dimension 16 and then concatenated with the input
going into a final dense layer of size 6 with a softmax activation
function. Dropout layers of $0.5$ are applied after the input and
before the output layers. 

The other 4 models are copies of the architecture described in \ref{subsec:Mini-ASC-Modules}.
All orange layers of size 6 are outputs of the model and are trained
against the labels with equal contribution to the total loss. We used
the Adam optimizer with a batch size of 32, a learning rate of $5\cdot10^{-4}$
and a decay of $5\cdot10^{-5}$ (decay in Adam is introduced in Keras,
and is not part of the original algorithm; it represents decay between
batches). This network reaches an average F\textsubscript{1} score
of $0.700$ on the validation set. This first ensemble is denoted
by E1.

\subsection{Second Ensemble\label{sec:Ensemble}}

In the second level ensemble (figure 
\begin{figure*}
\begin{centering}
\includegraphics[scale=0.5]{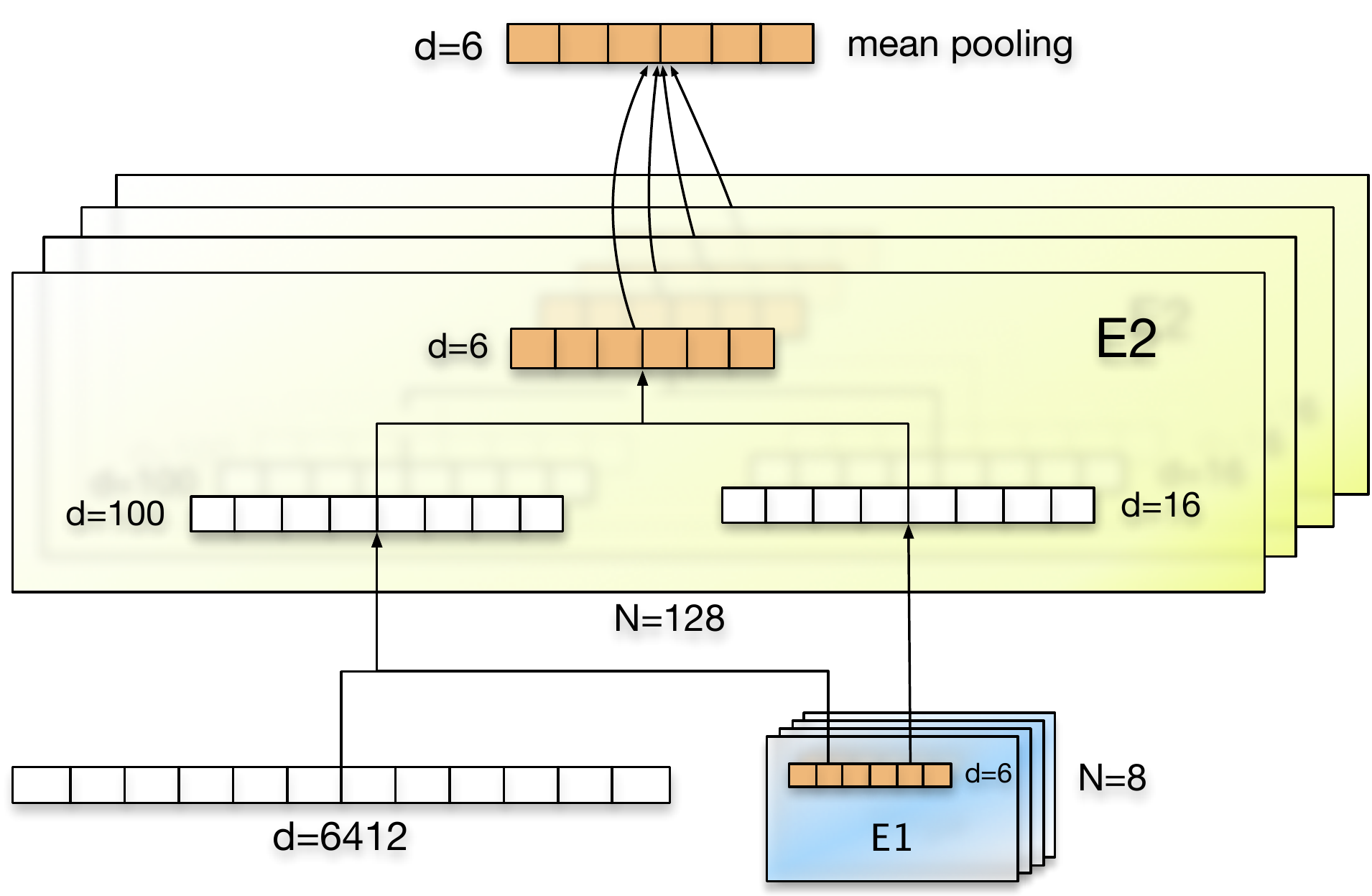}
\par\end{centering}
\caption{\label{fig:ensemble}Architecture of the second-level ensemble.}
\end{figure*}
\ref{fig:ensemble}), we started with 8 copies of the aforementioned
E1 models (with different parameters for the Mini ASC modules in the
ranges described in section \ref{subsec:Mini-ASC-Modules}) and combined
them with a concatenation of the following features (described in
section \ref{sec:Features}): two external embeddings (Universal Sentence
Embedding, DeepMoji) and our Semeval 2018 pipeline predictions.

We have used a dense layer of size 16 over the outputs of the 8 E1
models and a dense layer of size 100 over all of the above features
(including the E1 outputs). These two layers were concatenated into
a softmax layer of size 6 which was the output of the second ensemble;
we denote this by E2. This E2 network reaches an average F\textsubscript{1}
score of 0.702 on the validation set. The final model is a soft voting
ensemble, comprising 128 networks of type E2; this probability averaging
is meant to decrease the variance of the model which reaches an average
F\textsubscript{1} score of 0.705 on the validation set. 

Since the final model is an ensemble, where some models are somewhat
overfitted with respect to the training dataset (e.g. E1) and some
models are not overfitted at all (LM1), we decided to use the validation
dataset to train the final model for an additional 4 epochs using
a large batch size of 900. After this procedure, the system scored
an F\textsubscript{1} of 0.7145 on the test dataset.

\section{Review and Conclusions\label{sec:Summary}}

In this paper we described the system developed for the WASSA 2018
implicit emotion shared task. It consists of a multi-level ensemble,
combining a novel use of language models to predict the right emotion
word, together with previous high-ranking architecture, used in the
Semeval 2018 sentiment shared task, and two external embeddings. The
system reached 1st place with macro F\textsubscript{1} of 0.7145,
with the next system scoring 0.7105. Examining the nature of the this
task, it is a combination of both sentiment classification and word
prediction; this was the motivation of using the Semeval 2018 models,
which were designed to classify emotions. On the other hand, the language
model is specifically trained to maximize the likelihood of matching
a word to a given sentence, thus naturally lending itself to the word
prediction aspect of the task. 

We have seen that splitting the dataset into two parts, one for training
our models and the other for the ensembling process (in this case
the second part is the validation set) is much more beneficial than
training our models on the combined bigger dataset, in cases when
some of the models are expected to be much less generalizable than
others. 

It is interesting to note the task organizers have tested human performance
on a subset sample, achieving macro F\textsubscript{1} of 0.45, which
is much lower than the automated systems.

We released the word embeddings\footnote{\url{https://s3.amazonaws.com/amobee-research-public/embedding/embedding.zip}} and language model\footnote{\url{https://s3.amazonaws.com/amobee-research-public/language-model/language-model.zip}} as open-source in order to benefit further research and increase sharing of resources.

\bibliographystyle{acl_natbib_nourl}
\bibliography{semeval2018}

\appendix
\onecolumn

\section{Emotions Lexicon\label{sec:Emotions-Lexicon}}

\begin{table}[h]
\begin{centering}
\begin{tabular}{cl}
\toprule 
Emotion & \multicolumn{1}{c}{Words}\tabularnewline
\cmidrule[0.05em](r){1-1}\cmidrule[0.05em](l){2-2}Anger & Anger, angry, fuming, angrily, angrier, angers, angered, furious.\tabularnewline
Disgust & Disgust, disgusted, disgusting, disgustedly, disgusts.\tabularnewline
Fear & Fear, feared, fearing, fearfully, frightens, fearful, afraid, scared.\tabularnewline
Joy & Joy, happy, thrilling, joyfully, happily, happier, delights, joyful,
joyous.\tabularnewline
Sad & Sad, sadden, depressing, depressingly, sadder, depresses, sorrowful,
saddened.\tabularnewline
Surprise & Surprise, surprised, surprising, surprisingly, surprises, shocked.\tabularnewline
\bottomrule
\end{tabular}
\par\end{centering}
\caption{\label{tab:Emotion-lexicon}Emotion lexicon used to produce predictions
using the language models.}

\end{table}

\end{document}